\documentclass[journal]{IEEEtran}

\usepackage{xcolor,soul,framed} 

\colorlet{shadecolor}{yellow}
\usepackage[pdftex]{graphicx}
\graphicspath{{../pdf/}{../jpeg/}}
\DeclareGraphicsExtensions{.pdf,.jpeg,.png}

\usepackage{cite}
\usepackage[cmex10]{amsmath}
\usepackage{bbm}
\usepackage{multirow}
\usepackage{array}
\usepackage{mdwmath}
\usepackage{mdwtab}
\usepackage{eqparbox}
\usepackage{url}
\usepackage{booktabs}
\usepackage{caption}
\usepackage{hyperref}
\hypersetup{
    colorlinks=true,
    linkcolor=blue,
    filecolor=magenta,      
    urlcolor=cyan
}
    
\usepackage[switch]{lineno}
    
\begin{document}
\bstctlcite{IEEEexample:BSTcontrol}
\title{Conditional Predictive Behavior Planning with Inverse Reinforcement Learning for Human-like Autonomous Driving}
\author{Zhiyu Huang,
        Haochen Liu,\
        Jingda Wu,\
        and Chen Lv,~\IEEEmembership{Senior Member,~IEEE}

\thanks{Z. Huang, H. Liu, J. Wu, and C. Lv are with the School of Mechanical and Aerospace Engineering, Nanyang Technological University, Singapore, 639798. (E-mails: zhiyu001@e.ntu.edu.sg, haochen002@e.ntu.edu.sg, jingda001@e.ntu.edu.sg, lyuchen@ntu.edu.sg).}%
\thanks{This work was supported in part by the SUG-NAP Grant, Nanyang Technological University, and the A*STAR AME Young Individual Research Grant, Singapore (No. A2084c0156).}%
\thanks{Corresponding author: C. Lv.}}


\maketitle

\begin{abstract}
Making safe and human-like decisions is an essential capability of autonomous driving systems, and learning-based behavior planning presents a promising pathway toward achieving this objective. Distinguished from existing learning-based methods that directly output decisions, this work introduces a predictive behavior planning framework that learns to predict and evaluate from human driving data. This framework consists of three components: a behavior generation module that produces a diverse set of candidate behaviors in the form of trajectory proposals, a conditional motion prediction network that predicts future trajectories of other agents based on each proposal, and a scoring module that evaluates the candidate plans using maximum entropy inverse reinforcement learning (IRL). We validate the proposed framework on a large-scale real-world urban driving dataset through comprehensive experiments. The results show that the conditional prediction model can predict distinct and reasonable future trajectories given different trajectory proposals and the IRL-based scoring module can select plans that are close to human driving. The proposed framework outperforms other baseline methods in terms of similarity to human driving trajectories. Additionally, we find that the conditional prediction model improves both prediction and planning performance compared to the non-conditional model. Lastly, we note that the learning of the scoring module is crucial for aligning the evaluations with human drivers.
\end{abstract}


%
\IEEEpeerreviewmaketitle

\section{Introduction}
\IEEEPARstart{M}{aking} human-like decisions is crucial for autonomous vehicles (AVs) because they need to operate among humans in a safe and socially-compliant manner. The modern autonomous driving stack divides the decision-making task into two sub-modules: behavior planning and trajectory planning \cite{sadat2019jointly}. Behavior planning determines high-level decisions, such as lane changes, overtaking, and yielding, while trajectory planning generates a smooth trajectory to achieve these high-level objectives. Behavior planning plays a paramount role in the decision-making process because it informs the downstream trajectory planning and directs the final control outputs. However, widely used behavior planners rely on hand-engineered rules or finite-state machines to specify the vehicle's behaviors under different situations \cite{van2020hierarchical, censi2019liability}. However, this approach is not scalable as the rules are difficult to maintain and can conflict with each other as more scenarios are considered. Additionally, for complex real-world scenarios such as intersections with a complex road structure and many different types of traffic participants, it is challenging to design rules that align with human expectations. To address these limitations, end-to-end learning-based methods (e.g., imitation learning and reinforcement learning) that directly make decisions from perception results or raw sensors have gained popularity. However, these methods lack reliability, interpretability, and safety guarantee. On the contrary, this work proposes a predictive behavior planning framework \cite{wei2014behavioral} that predicts the behaviors of surrounding road users and evaluates the goodness of synthesized decisions using a cost function explicitly considering speed, comfort, and safety. The framework is believed to be closer to the human decision-making process, making it more scalable, robust, and suitable for real-world scenarios. Our framework aims to address the scalability issue by learning from a variety of real-world scenarios and making decisions that are closer to human driving. The main metric used to measure human likeness is the displacement error compared to human driving trajectories.

In this paper, we focus on learning the prediction model and evaluation model in the framework, and we try to address two specific challenges. The first challenge is to improve the prediction accuracy and make the prediction results more suitable for the downstream task. Deep learning-based approaches \cite{liang2020learning, huang2021multi, gu2021densetnt} have shown significant improvement in accuracy compared to conventional ones (e.g., kinematic models and intelligent driver model). However, most motion prediction models are only passively used in behavior planning, meaning the prediction model often outputs fixed results for other agents and ignores the potential impact of the ego vehicle's future actions. This can often lead to overly conservative and even dangerous decisions. To overcome this issue, we leverage the conditional motion prediction (CMP) method \cite{tolstaya2021identifying, ngiam2021scene, tang2022interventional} in our behavior planning framework, which jointly predicts the future motions of multiple interacting agents in a scene based on the AV's candidate decisions. This provides the evaluation module with more accurate information, enabling better evaluation of the consequences of different decisions. The second challenge is to design the cost function, which determines which behaviors are desirable, because there are many hard-to-specify nuances in human driving behaviors (e.g., speed profiles, comfort, and preference of risk). Manually tuning the cost function can be laborious and may not reflect the actual human preferences, resulting in unintended behaviors. To resolve this, we employ a maximum entropy inverse reinforcement learning (IRL) framework \cite{ziebart2008maximum} to automatically learn the cost function from human driving data.

Our proposed conditional predictive behavior planning framework consists of three main components: generation, prediction, and scoring. The trajectory \textbf{generation} module uses a performant and interpretable approach to create diverse trajectory proposals for the AV considering factors such as lane, traffic rules, and speed profiles. The conditional motion \textbf{prediction} (CMP) module then forecasts the surrounding agents' future trajectories based on each trajectory proposal and yields different scene-compliant trajectories. The CMP module in our framework is inspired by \cite{song2020pip}, but we propose a novel Transformer-based structure in observation encoding and plan fusion. The prediction network is built upon our previous work \cite{huang2022differentiable}, however, the influence of the {AV’s} future actions is only implicitly considered in training time in our previous work, while we explicitly consider it in test time in this work. The \textbf{scoring} module with a learnable cost function evaluates these trajectory proposals to make final decisions. Specifically, the factors we consider in the cost function encompass travel efficiency (speed), ride comfort (longitudinal jerk and acceleration and lateral acceleration), risk (headway and lateral distance), as well as safety (collision probabilities with other vehicles). Different from our previous work on IRL \cite{huang2021driving} which uses either a perfect environment model (log replay) or a simple one, we use the proposed conditional prediction network as the environment model, leading to better performance in test time. The main contributions of this paper are summarized as follows:
\begin{enumerate}
\item We propose a learning-based behavior planning framework that learns to predict conditional multi-agent future trajectories and evaluate decisions from real-world human driving data.
\item We propose a novel Transformer-based conditional motion prediction model that utilizes the attention mechanism for effective environment encoding and plan fusion.
\item We propose a two-stage learning process where the conditional prediction model is trained first and used as an environment model in the learning of the cost function with maximum entropy IRL.
\item We conduct comprehensive experiments to demonstrate the ability of our framework to make conditional and multi-modal predictions and human-like decisions.
\end{enumerate}

\begin{figure*}[htp]
    \centering
    \includegraphics[width=0.95\linewidth]{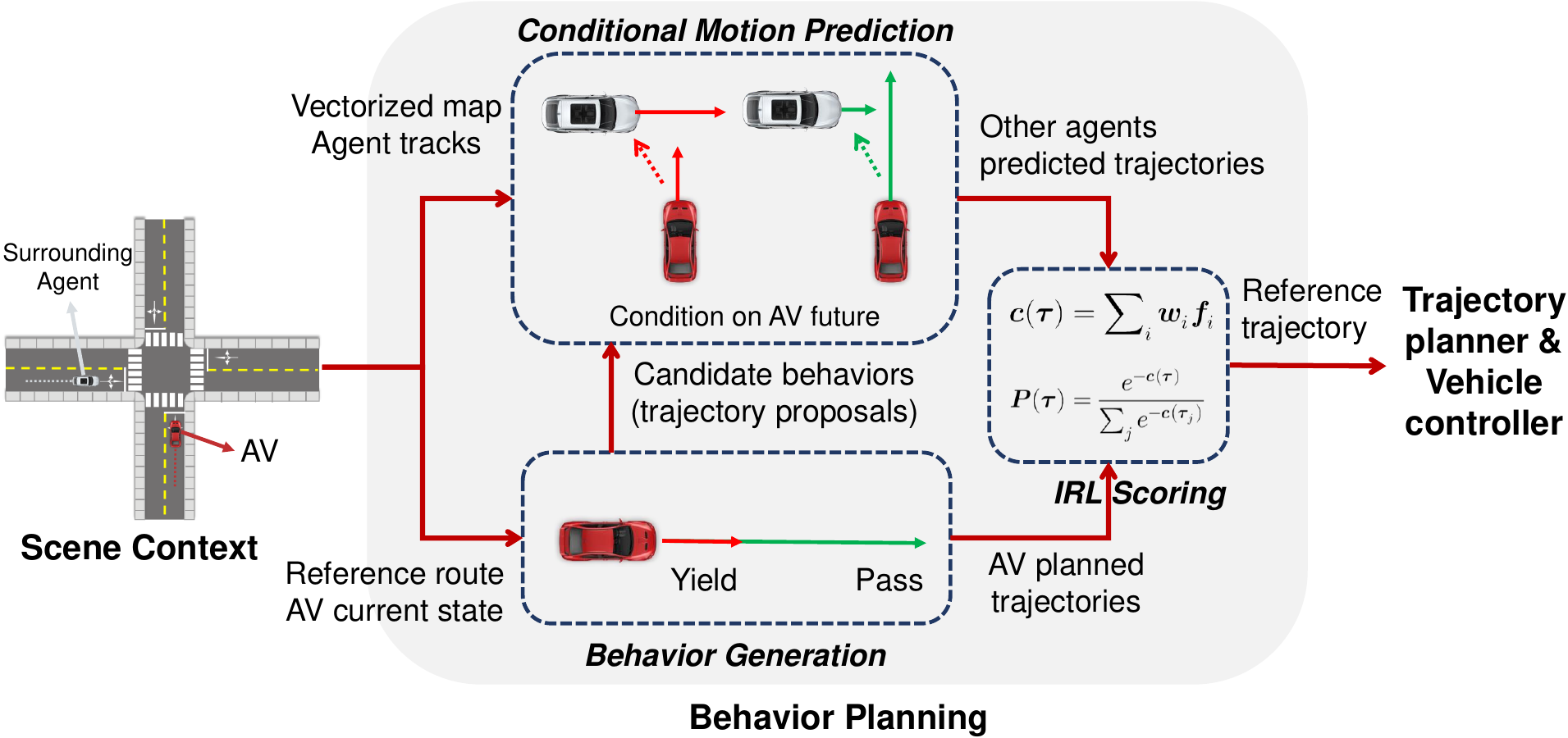}
    \caption{The proposed behavior planning framework. The behavior generation module synthesizes a set of candidate behaviors. The conditional motion prediction module takes as input the vectorized map, agent tracks, and the AV's future plan to predict other agents' future trajectories. The IRL scoring module calculates the probability of each behavior based on the features of joint trajectories of the AV and other agents.}
    \label{fig:1}
\end{figure*}

\section{Related Work}
\label{sec2}

\subsection{Decision-making for autonomous driving}
In addition to perception ability \cite{9945651}, decision-making has become a research hotspot for autonomous driving in recent years due to its importance as a bottleneck for the widespread deployment of autonomous vehicles. In particular, learning-based decision-making methods such as imitation learning (IL) and reinforcement learning (RL) have shown promising results and potential \cite{9910009}. IL attempts to directly imitate the actions of a human driver through the use of holistic neural networks and massive offline driving datasets \cite{bansal2018chauffeurnet, huang2020multi, chitta2021neat}. However, an evident issue with IL is that there is a distribution shift from training to deployment, which is difficult to mitigate. On the other hand, RL tries to optimize a reward function by interacting with the environment and learning from trial and error, but still faces obstacles such as sample efficiency, accurate environment modeling, and proper reward function design \cite{liu2022improved, wu2022uncertainty, wu2022toward, liu2022augmenting, huang2022efficient, wu2022prioritized}. Despite these ongoing efforts on learning-based methods, IL and RL have their inherent flaws, and a fundamental problem is that their safety, interpretability, and generalizability are somehow compromised. As a result, there has been a shift towards classic planning methods (e.g., graph search, sampling, and optimization) \cite{gonzalez2015review}, which provide stronger safety guarantees, rule compliance, and interpretability. Nevertheless, their performance heavily relies on the prediction accuracy of surrounding agents and proper evaluation of the planned behaviors, which are the main focuses of this work.

\subsection{Motion prediction}
There has been a growing body of deep learning-based motion prediction approaches thanks to the wide availability of large-scale driving datasets \cite{huang2022recoat, mo2022stochastic}. They have demonstrated excellent accuracy and scalability because of their ability to handle high-dimensional map data and model agent interactions. \cite{gao2020vectornet} proposed a vectorized high-definition map representation, which is more compact and easy to process than an image-based map. To model the agent-agent and agent-map interactions, the agents and map vectors are often abstracted into a graph and processed by graph neural networks (GNNs) \cite{gilles2021gohome, mo2022multi} or Transformer networks \cite{huang2021multi, ngiam2021scene}. In addition, the prediction model should be capable of outputting multiple possible futures (to address uncertainty) for multiple surrounding agents in the scene (for efficient inference and better scene consistency) \cite{ngiam2021scene, su2022narrowing}.

However, most of the motion prediction models ignore the influence of the AV's future actions on other agents, and the decision-making module has to act passively. This is problematic because other agents may react differently if the AV makes different decisions, and ignoring this may result in the AV's too conservative or aggressive behaviors. More recently, conditional motion prediction (CMP) has emerged to address this issue and enable more interactive prediction. \cite{tolstaya2021identifying} proposed a CMP model that predicts future trajectories for other agents conditioned on a query future trajectory for an ego agent, which shows a $10$\% improvement in accuracy over non-conditional prediction. M2I \cite{sun2022m2i} extended CMP to multi-agent interactive prediction, where an influencer is selected and reactors' future trajectories are predicted using a CMP model according to the influencer's marginal prediction result. Scene Transformer \cite{ngiam2021scene} proposed a unified Transformer-based architecture with a masking strategy, enabling one to predict other agents' behaviors conditioned on the future trajectory of the AV. However, these works still focus on the prediction task, and CMP models are not integrated into planning. PiP \cite{song2020pip}, which conditions the prediction process on multiple candidate future trajectories of the AV, is the most related to our work. However, the planning performance was not investigated, probably due to the difficulty in evaluating the candidate trajectories, and the model is only validated on highway datasets. Our method is more focused on improving the downstream planning performance and validated on more challenging urban driving datasets.

\subsection{Inverse reinforcement learning}
Inverse reinforcement learning (IRL) methods aim to learn underlying cost functions from expert demonstrations, thereby avoiding manual specification. Maximum entropy IRL approach \cite{ziebart2008maximum}, which addresses ambiguities or uncertainties inherent in human demonstrations, has become popular in autonomous driving applications. \cite{kuderer2015learning} applied maximum entropy IRL to learn individual driving styles from highway driving demonstrations and reproduce distinct driving policies. \cite{rosbach2019driving} combined maximum entropy IRL with a planning algorithm to automatically tune the cost function, which exceeds the level of manual expert-tuned cost functions. However, they either ignored other agents (which is not practical) or used very simple models (e.g., constant speed) to predict other agents' actions. In our previous work using sampling-based IRL to infer cost functions for human drivers \cite{huang2021driving}, we build a simple environment model that uses log replay and an intelligent driver model to simulate other agents' actions when they are influenced. In this work, we improve the prediction module to a neural network-based conditional prediction model and integrated it into a decision-making module based on IRL scoring, which allows the framework extends to complex urban driving scenarios.


\section{Method}
The proposed behavior planning framework is illustrated in Fig. \ref{fig:1} and consists of three core components: behavior generation, conditional motion prediction, and IRL scoring. The behavior generation component synthesizes a diverse set of trajectory proposals by sparsely sampling plausible maneuvers and velocity profiles, based on the current state and reference route of the AV. The conditional motion prediction component predicts the trajectories of surrounding agents under each planned trajectory, providing multiple possible futures for both the AV and other agents. The features of the different trajectory proposals (including cumulative step-wise features over the time horizon and trajectory-level features) are combined with learnable weights to calculate the costs, and the probabilities of trajectory proposals are obtained according to maximum entropy IRL. The behavior (trajectory proposal) with the highest probability or sampled from the distribution can be used as a reference trajectory in the downstream trajectory planning, which further refines the coarse trajectory and generates a well-defined trajectory for the autonomous vehicle to follow. In this work, we omit the trajectory planning module since it is well-established. In the following, we will first formulate the behavior planning problem and then elaborate on the key components of our framework. 

\subsection{Problem formulation}
Consider an arbitrary driving scene consisting of $N+1$ agents, where the AV denoted as $A_0$ and other surrounding agents denoted as $A_i \ (i=1, \cdots, N)$, and let $\mathbf{s}_t^i$ represent the physical state of agent $i$ at time step $t$. Assuming the current time step $t=0$, $\mathbf{S}^i_{-T_h:0} \doteq \{ \mathbf{s}_{-T_h}^i, \cdots, \mathbf{s}_0^i \}$ denotes the historical states over a past horizon of length $T_h$. Given the joint historical states of surrounding agents $\mathbf{X} \doteq \mathbf{S}^{1:N}_{-T_h:0}$, as well as the map information $\mathcal{M}$, the conventional motion prediction task is to model the posterior distribution $p(\mathbf{Y}|\mathbf{X}, \mathcal{M})$, where $\mathbf{Y} \doteq \mathbf{\hat S}^{1:N}_{1:T_f}$ denotes the joint states of surrounding agents over a future time horizon $T_f$. For conditional motion prediction, we incorporate additional information on the AV's potential future trajectory $\mathbf{P} \doteq \{ \mathbf{\hat s}_{1}^0, \cdots, \mathbf{\hat s}_{T_f}^0 \}$, which is to model the conditional distribution $p(\mathbf{Y}|\mathbf{X}, \mathcal{M}, \mathbf{P})$.

Given the planning and conditional prediction results $\mathbf{P}$ and $p(\mathbf{Y}|\mathbf{P})$, we utilize a cost function $f$ to score the results and make decisions. In particular, let $r(\mathbf{P})$ represent the reward (negative cost) of the planned trajectory and $r(\mathbf{P}) = - f(\mathbf{P}, p(\mathbf{Y|\mathbf{P}}); \mathbf{w})$, where $\mathbf{w}$ is the learnable weights. The probability of a planned trajectory being selected according to the maximum entropy principle is $P(\mathbf{P}_{i}) = \frac{\exp r(\mathbf{P}_{i})}{\sum_{j} \exp r(\mathbf{P}_{j})}$. We adopt the IRL algorithm to learn the weights $\mathbf{w}$ from expert demonstrations, which is to maximize the log-likelihood of the expert demonstration trajectory $\mathbf{P}^*$ being selected: $\max_{\mathbf{w}} \log P(\mathbf{P}^*|\mathbf{w})$.

\subsection{Behavior generation}
We generate different candidate behaviors for the AV in the format of trajectory proposals. This is done by considering the road structure (available lane centerlines), semantics (changing lanes and lane-keeping, as well as different speed profiles), and traffic rules (speed limit). The trajectory proposals are generated by utilizing polynomial curves given the target lane and speed. The process ensures that the generated trajectories are dynamically feasible and the behaviors are correct, interpretable, and comply with the traffic rules. To adapt to complex road structures in urban scenarios, we represent a trajectory in the Frenet Frame of the reference path (e.g., the lane centerline) \cite{werling2010optimal} and then translate it back to the Cartesian coordinate. We consider a fixed time horizon $T=5s$ because it can well balance prediction accuracy and long-term decisions and satisfy the requirement of generating coarse trajectory proposals. Specifically, for the longitudinal $s$-axis, we specify a set of target velocities at the terminal state ranging from braking to stop ($\dot s(T)=0$) to accelerating to the speed limit ($\dot s(T)=v_{limit}$), while the target acceleration is fixed to be $0$ ($\ddot s(T)=0$). We use a quartic polynomial to parameterize the longitudinal states along the trajectory:
\begin{equation}
\label{eq1}
s(\tau) = a_0 + a_1\tau + a_2\tau^2 + a_3\tau^3 + a_4\tau^4  
\end{equation}
where $\tau$ is the time, $\{a_0, \cdots, a_4\}$ are the coefficients. Given the initial state $[s(0), \dot s(0), \ddot s(0)]$  and target state $[\dot s(T), \ddot s(T)]$, we can calculate the coefficients and get the longitudinal state at each time step on the trajectory.

For the lateral $d$-axis, we assume the AV finishes the maneuver at the end of the time horizon, i.e., $d(T)=0$, $d(T)=D$, or $d(T)=-D$, where $D$ is the distance to the neighbor lanes, and fix the end conditions $\dot d(T)$ and $\ddot d(T)$ to be 0. We use a quintic polynomial to represent the lateral states: 
\begin{equation}
\label{eq2}
l(\tau) = b_0 + b_1\tau + b_2\tau^2 + b_3\tau^3 + b_4\tau^4  + b_5\tau^5    
\end{equation}
where $\{b_0, \cdots, b_5\}$ are the coefficients of the polynomial. Likewise, given the initial state $[d(0), \dot d(0), \ddot d(0)]$ and target state $[d(T), \dot d(T), \ddot d(T)]$, we can calculate the coefficients and obtain the lateral states.

\begin{figure}[htp]
    \centering
    \includegraphics[width=\linewidth]{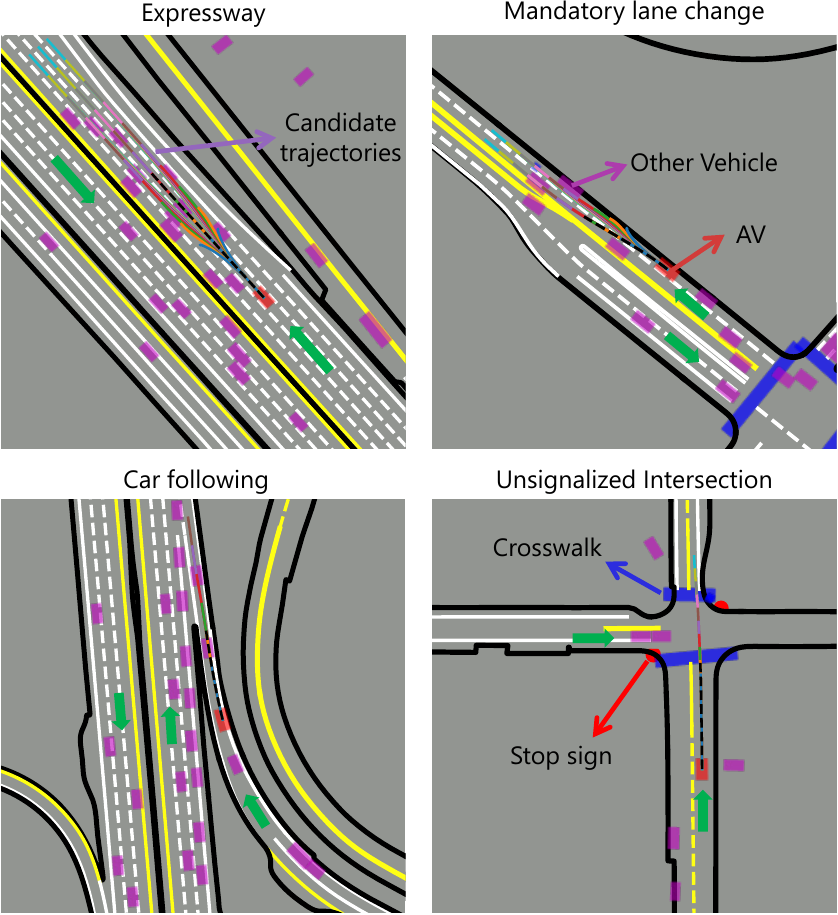}
    \caption{Representative examples of the behavior generation process in a variety of urban driving scenarios. The colored lines are generated candidate trajectories and the black dotted lines are ground-truth trajectories.}
    \label{fig:2}
    \vspace{-0.5cm}
\end{figure}

The final trajectory is represented by a sequence of states with regard to the $s$ and $d$ axis $[s(\tau), d(\tau)]$, which is then translated back to the Cartesian coordinate $[x(\tau), y(\tau)]$ given the reference {route’s} coordinate, heading angle, and curvature \cite{werling2010optimal}. Approximately $10-30$ candidate behaviors (trajectory proposals) are generated depending on the AV's initial state and road structure. Fig. \ref{fig:2} shows some representative cases of the module generating candidate behaviors for the AV according to different road structures. The examples provided demonstrate the capabilities of our method in handling various scenarios in urban driving. Given the candidate lane centerlines, our proposed method is capable of addressing nearly all scenarios on structured roads. More details about the behavior generation process can be found in section \ref{trajectory}.

\subsection{Conditional motion prediction}
\begin{figure*}[htp]
    \centering
    \includegraphics[width=\linewidth]{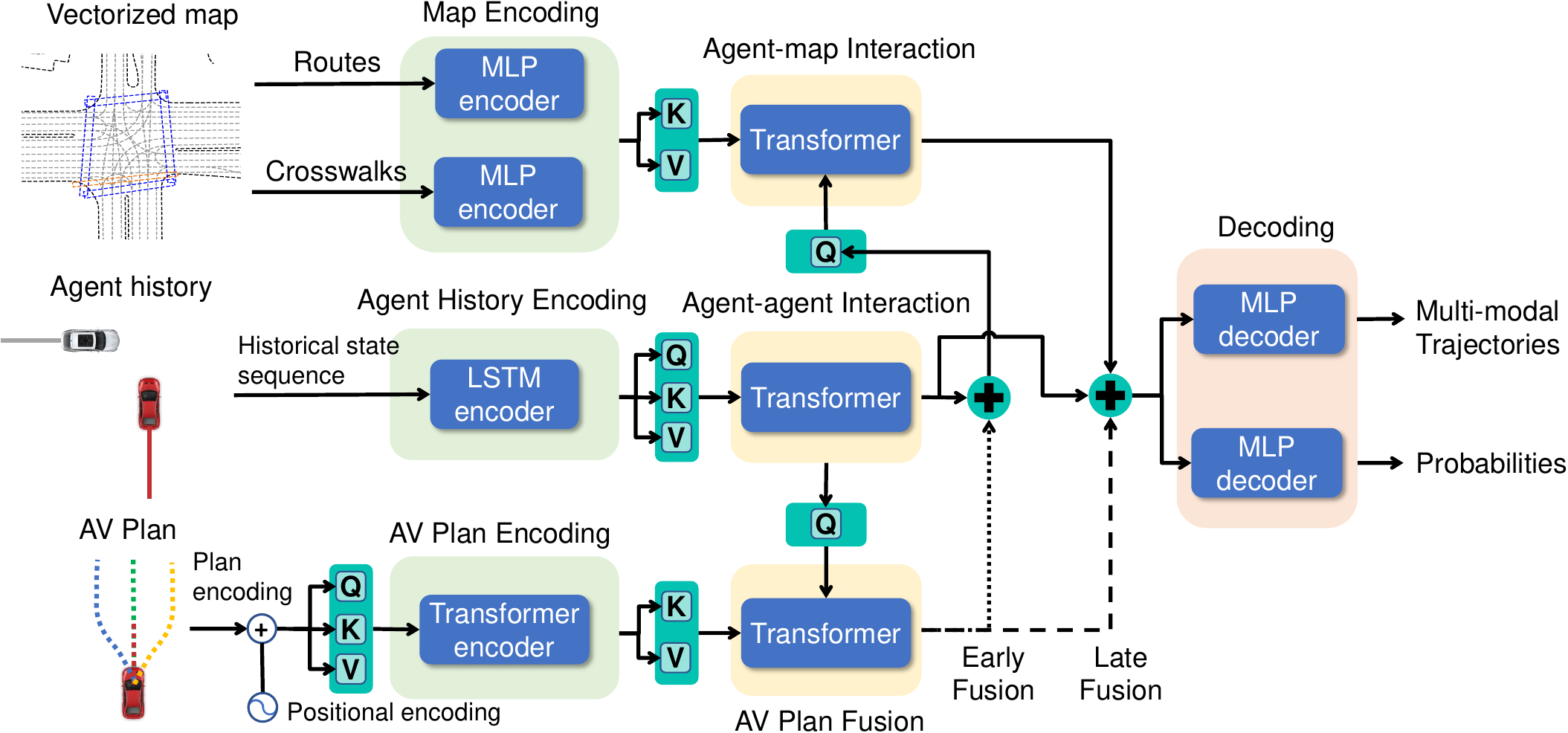}
    \caption{The structure of the conditional motion prediction network. The information on the AV's future plan is fused into the network to realize that the predictions of other agents are conditioned on the AV's planned trajectory.}
    \label{fig:3}
\end{figure*}

We utilize a conditional motion prediction module to predict other agents' future motions conditioned on the AV's planned trajectory. The prediction module needs to address the inherent uncertainties (multi-modality) of other agents' intentions, and thereby predict a diverse set of future outcomes for other agents. Particularly, we represent the future outcomes as Gaussian mixture model (GMM), where each mixture component corresponds to a joint future state sequence for all other agents. The state of a multi-agent joint future at each time step is represented as a Gaussian distribution:
\begin{equation}
\phi (\mathbf{\hat S}^{1:N}_{t}|\mathbf{P}) = \mathcal{N} (\mu_t(\mathbf{P}), \Sigma_t(\mathbf{P})),
\end{equation}
where $\mu_t$ and $\Sigma_t$ are the mean and covariance, respectively. The agents' historical states $\mathbf{X}$ and map information $\mathcal{M}$ are omitted for conciseness.

We model a probabilistic distribution (GMM) over multiple joint future outcomes, which represents the probability over each predicted future: 
\begin{equation}
p(\mathbf{Y}|\mathbf{P}) = \sum_{k=1}^K p (\mathbf{Y}^k|\mathbf{P}) \prod_{t=1}^{T_f} \phi (\mathbf{\hat S}^{1:N}_{t}|\mathbf{P}, \mathbf{Y}^k),
\end{equation}
where $K$ is the number of the mixture components, $p (\mathbf{Y}^k)$ is the probability of the $k$-th component. 

The multi-step means and covariances of each future, as well as its probability, are learned parameters. We design a Transformer-based neural network based on our previous work \cite{huang2022differentiable} to fulfill the conditional prediction task. The architecture of the prediction network is shown in Fig. \ref{fig:3} and the details of the building blocks are described below.

\textbf{Input Representations}.
The prediction network utilizes three diverse sources of information: vectorized map, agent history, and AV plan. In practice, these kinds of information are readily available in a modern autonomous driving system. The agent history tracks can be obtained by the object detection and tracking function in the perception system, and the vectorized map can either be obtained from a high-definition map created offline or from an onboard mapping system. The candidate plans for the AV are generated using the proposed method. For each agent, we find its possible driving routes stretching a predefined length ($100$ meters) and extract the waypoints with a fixed interval ($1$ meter), as well as its nearby crosswalk polylines.

\textbf{Agent History and Map Encoding}. 
The historical states of all agents (including the AV) are encoded by different dual-layer long short-term memory (LSTM) networks for different types of agents (i.e., vehicle, pedestrian, and cyclist). For each agent, we apply two multi-layer perceptrons (MLPs) to encode the driving routes and crosswalks respectively. A map waypoint is comprised of spatial attributes, such as its position and heading angle, as well as additional attributes like speed limit and traffic signals, which are effectively encoded by the MLP as the latent feature of that waypoint. More details about the encoders can be found in \cite{huang2022differentiable}.

\textbf{AV Plan Encoding}. 
A candidate behavior is represented as a trajectory proposal, which is filled into a tensor with shape $(T_f, D_p)$. The future state $D_p$ of a planned trajectory consists of the $x$ and $y$ coordinates, heading angles, and speed. Specifically, the trajectory is projected to a high dimension with an MLP and the positional encoding is added, and then we employ a self-attention Transformer encoder layer \cite{vaswani2017attention} to extract the temporal relation of the trajectory, obtaining the encoding of a planned trajectory with shape $(T_f, D)$. The same encoder is applied to all trajectory proposals.

\textbf{Interaction Modeling}. 
The agent-agent interaction is modeled by a two-layer self-attention Transformer encoder, which takes as input the agents' historical state encoding and outputs the feature of relations between them, and we remove the feature of the AV and obtain the agent interaction encoding with shape $(N, D)$. For each agent in the interaction encoding, we propose an agent-map encoder (two cross-attention layers and a multi-modal attention layer \cite{huang2021multi}) to generate different modes of agent-map interaction features with shape $(K, D)$. All the surrounding agents share the same agent-map interaction encoder. More details about the interaction modeling encoders can be found in \cite{huang2022differentiable}. 

\textbf{AV Plan Fusion}. Injecting the AV's planned trajectory as conditional information into other agents' encoded feature to predict their future trajectories is an essential part of the model. To fuse that information, we design two viable approaches, namely early fusion and late fusion. We first use a cross-attention Transformer layer with the agent-agent interaction features as query and AV plan encoding as key and value and obtain a tensor with shape $(N, D)$ representing the influence of the AV's future plan on different agents. In the early fusion approach, an attention-based fusion mechanism is designed by combining the agent-agent interaction feature and AV's future influence as the query to the agent-map interaction module. In contrast, the late fusion approach simply concatenates the feature of the AV's future plan with the final encoding, which is used to decode the future trajectories. We have also combined early fusion and late fusion (denoted as early + late fusion), and the effectiveness of the different fusion methods will be evaluated through experiments.

\textbf{Decoding}. We repeat the agent interaction encoding $K$ times along the zeroth dimension and concatenate it with the agent-map interaction encoding of all agents $(K, N, D)$, to generate a final latent representation tensor with shape $(K, N, 2D)$. It contains the necessary information to predict an agent's future motion, including its historical physical states, interaction with other agents, relation with the map, and the influence of the AV's plan if using early fusion. For late fusion, the influence of the AV's future plan on other agents is also concatenated to the latent representation tensor, yielding a tensor with shape $(K, N, 3D)$. We use an MLP to decode the Gaussian parameters $\mathcal{N} (\mu_x, \sigma_x; \mu_y, \sigma_y)$ for every agent at every timestep in the future from the latent representation, outputting a tensor with shape $(K, N, T_f, 4)$. To predict the probability of different futures, which is a tensor with shape $(K, 1)$, we first use max-pooling to aggregate the information of all agents and then pass the obtained tensor with shape $(K, 2D)$ or $(K, 3D)$ through another MLP. 


\subsection{Inverse reinforcement learning}
Although we have obtained the candidate behaviors and other agents' predicted reactions through the CMP module, it is still challenging to appropriately evaluate the behaviors and make human-like decisions, because there are many hard-to-specify nuances in driving behaviors and different people may take different actions even in the same situation. Therefore, we adopt the maximum entropy IRL method \cite{ziebart2008maximum} to learn to evaluate these behaviors from human driving data.

Maximum entropy IRL aims to recover the underlying reward functions from demonstrations of human behaviors. It can address the ambiguity of multiple solutions and stochasticity of expert behaviors by recovering a distribution over all trajectories. In essence, according to the principle of maximum entropy, the resulting probability distribution over candidate behaviors (trajectories) is:
\begin{equation}
P(\zeta_i) = \frac{e^{-c(\zeta_i)}}{\sum_j e^{-c(\zeta_j)}},
\end{equation}
where $\zeta_i$ is a trajectory among the set of candidate trajectory proposals, and $c(\zeta_i)$ is the cost of that trajectory.
 
To maintain interpretability, we use a linear cost function, which is a linear combination of different features that characterize the driving behavior, and the cost of a candidate trajectory is the weighted sum of these features:
\begin{equation}
c(\zeta) = \mathbf{w}^\top \mathbf{f}(\zeta) = \sum_{i} w_i f_i(\zeta),   
\end{equation}
where $\mathbf{w}$ is the weights of the cost function and $\mathbf{f}(\xi)$ is the feature vector of the trajectory. The features of a trajectory are designed to cover the major concerns of autonomous driving, such as travel speed, ride comfort, traffic rules, and most importantly safety, which are detailed in section \ref{feature}.

The objective of IRL is to optimize the cost function weights in order to maximize the log-likelihood of the expert demonstration trajectories $\zeta^* \in D$ in the dataset:
\begin{equation}
\mathbf{w}^* = \arg \max_{\mathbf{w}} \sum_{\zeta^* \in D} \log P(\zeta^*|\mathbf{w}).
\end{equation}

We can use a gradient-based optimization method with automatic differentiation tools to optimize the weights.

\subsection{Learning process}
The learning process is divided into two stages. The first stage deals with conditional motion prediction, which is to learn to predict other agents' behaviors conditioned on the AV's future trajectory from a massive amount of interactions among human drivers. Since we can only get access to the ground-truth trajectories for the AV and other agents, we use the AV's ground-truth future trajectory as the planned trajectory and other agents' joint future trajectories are predicted and conditioned on the information. Although we cannot exactly learn how other agents would react to the AV's different plans in a specific scenario from an offline dataset, the CMP model is able to predict reactive and different behaviors of other agents given different plans. This is because the model can learn the reactions of other agents in a wide variety of scenarios where the AV takes different plans, and then generalize to predict other agents' reactions to different plans of the AV in a specific scenario.

To train the CMP module (a deep neural network parameterized by $\theta$), for each agent at a specific future time step, we adopt the negative log-likelihood (NLL) loss on the GMM parameters:
\begin{equation}
\begin{split}
& \mathcal{L}_{CMP} = - \log \mathcal{N}_{\hat k} (Y_x - \mu_x, \sigma_x; Y_y - \mu_y, \sigma_y; \rho=0) \\ 
& = \log \sigma_x \! + \log \sigma_y \! + \frac{1}{2} \! \left( \left(\frac{Y_x \! - \mu_x}{\sigma_x}\right)^2 \! + \left(\frac{Y_y \! - \mu_y}{\sigma_y}\right)^2 \right) \! - \log p_{\hat k},
\end{split}
\end{equation}
where $(Y_x, Y_y)$ represents the ground-truth position, $\hat k$ is the selected predicted future with the closest joint trajectories to the ground-truth ones measured by the L2 distance, $p_{\hat k}$ is the predicted probability of the selected Gaussian component. Note that we take the joint loss formulation by aggregating the losses of all agents jointly in the best-predicted future, in other words, all agents have the same best prediction mode $\hat k$. We adopt the cross entropy loss in the above equation to maximize the probability of the selected Gaussian component.

In the second stage, we concentrate on learning the cost function weights for evaluating the candidate plans. After we have obtained the well-trained CMP module, which outputs the possible trajectories of other agents given a planned trajectory, we can query the module for all the generated plans and get a set of future predictions. We then calculate the features and costs of all these futures (including features of the AV's trajectory and safety features considering other agents), and consequently the distribution of planned trajectories. We impose the NLL loss on the distribution, which favors the trajectory that most closely matches the expert demonstration in feature space:
\begin{equation}
\mathcal{L}_{IRL} = -\sum_{m=1}^M \mathbbm{1}(m=\hat m) \log P(\mathbf{P}_m|\mathbf{w}),
\end{equation}
where $\mathbf{P}_m$ is a planned trajectory, $\mathbf{w}$ is the learnable weights, $M$ is the number of generated plans, and $\hat m$ is the index of the plan with the closest end-point distance to the ground-truth trajectory.

\section{Experimental Validation}
\subsection{Dataset}
We train and validate the proposed framework (CMP and IRL modules) on the Waymo Open Motion Dataset (WOMD) \cite{Ettinger_2021_ICCV}, a large-scale real-world driving dataset, containing $104,000$ unique scenes (each $20$ seconds long at $10$ Hz) collected from $570$ hours of driving and over $1750$ km of roadways. The WOMD dataset provides annotated high-definition map data (e.g., lane polylines, lane connectivity, speed limits, and traffic signal states) and high-accuracy agent track data (e.g., coordinates, heading angles, velocities, and bounding box sizes), which is suitable for both prediction and planning tasks. 

In our experiments, we select $10,156$ scenes from the dataset and $80$\% of them are used as training data and the rest as testing data. In each scene, we split the 20-second long track data into several 7-second tracks with a sliding window, where the observation horizon is $2$ seconds and the prediction/planning horizon is $5$ seconds into the future. There is one track labeled as the self-driving car in each scene, and we choose it as the AV to perform behavior planning and its surrounding traffic participants are the agents to predict. We only utilize the AV track in each scene to train the IRL scoring module because they do not contain aggressive and unsafe behaviors. Eventually, we obtain $113,622$ training data points and $28,396$ testing data points. For the CMP module, all the data points are used for model training and evaluation of the prediction performance. For the IRL scoring module, we focus on the high-level behavior planning task and thereby filter those data points where the AV's average speed is less than $3 \ m/s$ (e.g., waiting at a red light), as well as a portion of data points where the AV cannot make lane changes. The amount of training data for the IRL scoring module after the filtering is $10,564$ and testing data for evaluating the planning performance is $2,246$.

\subsection{Behavior generation}
\label{trajectory}
To handle the complex road structures in urban areas, we generate candidate behaviors in the Frenet Frame of the given reference route so that we can separate the behaviors in the longitudinal and lateral directions. In the longitudinal direction, $10$ target speeds are evenly sampled in the range $[0, v_{limit}]$, where $v_{limit}$ is the speed limit of the route. In the lateral direction, according to the road structure, we specify the target lateral displacement to complete a lane change. For example, if there exists a left lane to the reference route that allows lane changing, we set the lateral displacement to be $D$, which is the distance from the current lane to the centerline of the left lane. The number of target lateral displacements varies with the road structure, from $1$ (keep lane) to $3$ (keep lane, change left, and change right). The different target speeds and lateral displacements are combined to generate candidate trajectories in Frenet space using Eq. (\ref{eq1}) and Eq. (\ref{eq2}). These trajectories are then translated back to Cartesian space and fed to the CMP module.

\subsection{Feature design}
\label{feature}
To maintain the interpretability of the planner, we design a set of representative features (scalar values) to characterize a candidate decision (trajectory). We can compute the features for each candidate trajectory and the cost of the trajectory is the sum of these features multiplied by the corresponding learnable cost weights: $c(\zeta) = \sum_i w_i f_i(\zeta)$. The designed features are described as follows.

\textbf{Travel efficiency}. 
We use the difference between the current speed and speed limit to represent the travel efficiency while also obeying traffic rules. The difference is normalized by the value of the speed limit to balance high-speed and the low-speed cases, and the feature takes the average value over all timesteps along the trajectory:
\begin{equation}
f_{\text{travel}}(\zeta) = \frac{1}{T_f} \sum_{t=1}^{T_f} \frac{|v_{t} - v_{limit}|}{v_{limit}},   
\end{equation}
where $v_t$ is the speed of the trajectory point at time step $t$, and $v_{limit}$ is the speed limit of the road.

\textbf{Maximum acceleration}. 
The maximum acceleration ($m/s^2)$ along the trajectory is used as a measurement of ride comfort, which is denoted as:
\begin{equation}
f_{\text{acc}} (\zeta) = \frac{\max_t |a_{t}^{lon}|}{a^{lon}_{\max}},
\end{equation}
where $a_{t}^{lon}$ is the longitudinal acceleration of the trajectory point at time step $t$, and $a^{lon}_{\max}=5 \ m/s^2$ is used to normalize this feature.

\textbf{Maximum jerk}. 
In addition to acceleration, we use the maximum jerk ($m/s^3)$ along the trajectory to represent the ride comfort in the longitudinal direction:
\begin{equation}
f_{\text{jerk}} (\zeta) = \frac{\max_t |j_{t}|}{j_{\max}},
\end{equation}
where $j_t$ is the longitudinal jerk of the trajectory point at time step $t$, and $j_{\max}=10 \ m/s^3$.

\textbf{Maximum lateral acceleration}.
The maximum lateral acceleration ($m/s^2)$ in the lateral direction is adopted as another measurement of ride comfort: 
\begin{equation}
f_{\text{lat\_acc}} (\zeta) = \frac{\max_t |a_{t}^{lat}|}{a^{lat}_{\max}},
\end{equation}
where $a_{t}^{lat}$ is the lateral acceleration of the trajectory point at time step $t$, and $a^{lat}_{\max}=5 \ m/s^2$.

The computation of the following features requires an estimate of other road users' states in the future, which is given by the CMP module. Note that we take the mean values of the predicted Gaussian at each timestep as a road user's state in calculating these features.

\textbf{Headway}.
The AV should keep a safe longitudinal distance to the leading vehicle, which is dependent on the speed of the AV. We utilize the concept of time headway and define the headway feature as follows.
\begin{equation}
\begin{aligned}
HW = \frac{1}{K} \sum_k & p_k \min_t \frac{\Delta s_t^{k}}{v_t},  \\
f_{\text{hw}} = & e^{-HW^2},
\end{aligned}
\end{equation}
where $\Delta s_t^{k}$ is the longitudinal distance between the AV and leading vehicle at timestep $t$ according to $k$-th predicted future, and $v_t$ is the speed of the AV at timestep $t$. We take the minimum value of time headway in the time horizon and average across all predicted futures by their probabilities. Then, we use a Gaussian radial basis function (RBF) to compute the headway feature, which aims to penalize the states the AV is too close to the leading vehicle.

\textbf{Lateral distance}.
The AV should keep a safe lateral distance from other vehicles, and thus we set up a feature to represent safety in the lateral direction.
\begin{equation}
\begin{aligned}
LD = \frac{1}{K} \sum_k & p_k \min_t \Delta l_t^{k},  \\
f_{\text{ld}} = & e^{-LD^2},
\end{aligned}
\end{equation}
where $\Delta l_t^{k}$ is the lateral distance between the AV and the closest vehicle on the sides of the AV at timestep $t$. Likewise, the minimum lateral distance in the time horizon from each predicted future is obtained and then weighed by the probability of the future. We also use a Gaussian RBF to compute the lateral distance feature, and if no other vehicles are on the sides, this feature is set to $0$. 

\textbf{Safety}.
The safety feature explicitly considers the collisions between the AV's planned trajectory and other vehicles' predicted trajectories, as well as uncertainties. At each time step, we calculate if the AV's planned state violates the spatial occupancy of any other agents in a given predicted trajectory. 
\begin{equation}
f_{\text{safety}} = \frac{1}{K} \sum_k p_k \sum_t \max_i \mathbbm{1}_{overlap}(\mathbf{\hat s}_t^{i, k}, \mathbf{\hat s}_t^0), 
\end{equation}
where $\mathbf{\hat s}_t^{i, k}$ is the predicted state of agent $i$ at timestep $t$ in the $k$-th future, $\mathbf{\hat s}_t^0$ is the state of the AV, and $\mathbbm{1}_{overlap}$ is an indicator function, which emits $1$ if the bounding box of the AV overlaps with an agent's bounding box and $0$ otherwise. If the AV collides with any other agents at timestep $t$, the frame is counted as a collision, and the collisions are summed across all time steps in the time horizon. The final safety feature averages the collision times from each predicted future weighted by their probabilities.

\subsection{Evaluation metrics}
To evaluate the prediction performance, two established metrics for behavior prediction are used, which are the minimum Average Distance Error (minADE) and minimum Final Distance Error (minFDE). minADE measures the average displacement of each point in the closest joint trajectories to the ground truth, while minFDE is the displacement error between the final point of the joint predicted trajectories and ground truth. The prediction errors are averaged for all agents in the joint trajectories.

We use a set of metrics to evaluate the behavior planning performance (primarily the closeness to human driving trajectories): minFDE between the top-3 most likely planned trajectories and ground-truth one, the accuracy of any of the top-3 most likely planned trajectories match with the ground-truth one, as well as the intention accuracy. We choose the top-3 accuracy because our planning framework is probabilistic, which can also address the stochasticity of human driving behaviors. In addition, we decrease the granularity of behaviors to discrete intentions, i.e., acceleration and deceleration in the longitudinal direction and lane change in the lateral direction, and we calculate the accuracy of our model to identify the intentions.

\subsection{Implementation details}
\label{details}
The parameters of the prediction module are listed in Table \ref{tab:1}. For all Transformer modules in the network, the number of attention heads is $8$, the hidden dimension of the feed-forward network is $1024$, and the activation function is RELU. Every dense layer except for the output layer is followed by a dropout layer with a dropout rate of $0.1$. The network outputs the displacements relative to an agent's current position instead of original coordinates, which could significantly improve prediction accuracy. We train the prediction module with the Adam optimizer and the learning rate with an initial value of 2e-4 decays by a factor of $0.5$ every $5$ epochs. The batch size is $32$ and the total training epochs is $30$. We clip the gradient norm of the network parameters with the max norm as $5$. 

\begin{table}[htp]
    \centering
    \caption{Parameters of the prediction module}
    \begin{tabular}{@{}ccc@{}}
    \toprule
    Symbol      & Meaning                                   & Value \\ \midrule
    $T_h$       & Length of historical timesteps            &  20     \\
    $T_f$       & Length of future  timesteps               &  50     \\
    $D_p$       & Dimension of AV plan features             &  4   \\
    $D$         & Dimension of embedding                    &  256   \\
    $N$         & Number of surrounding agents to consider  &  10   \\
    $K$         & Number of predicted futures               &  3     \\ \bottomrule
    \end{tabular}
    \label{tab:1}
\end{table}

For training the IRL-based planner (i.e., learning the cost function weights), we use the Adam optimizer with a learning rate that starts as 1e-2 and decays by a factor of $0.9$ every $50$ steps. We also add L2 regularization on the cost function weights with a weight decay value of 1e-2 to prevent overfitting. The mini-batch size is $64$ and the total training steps is $500$. To check if a collision happens between the AV and another object at a specific timestep, we approximate each object via a list of circles given their poses. The circles from the AV and the other object are paired, and if the distance between any pair of circles' centers is smaller than a threshold, it is considered that the two objects intersect and thus a collision happens. More details about the collision indicator can be found in \cite{scheel2022urban}.

\section{Results and Discussions}
\subsection{Prediction performance}
We first evaluate the performance of the conditional motion prediction module and report the results in both quantitative and qualitative manners. 

\begin{figure}[htp]
    \centering
    \includegraphics[width=\linewidth]{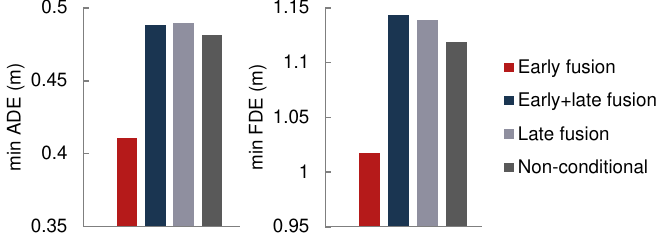}
    \caption{Quantitative results of the conditional prediction module with different fusion approaches}
    \label{fig:4}
\end{figure}

\textbf{Quantitative results}. 
Fig. \ref{fig:4} shows the quantitative prediction accuracy results in the testing set ($28,396$ scenes) for different network structures. Here, we use the ground-truth future trajectory of the AV as its planned trajectory and feed it to the CMP network. The structure of the non-conditional prediction model is the same as our proposed prediction model except that the AV plan encoding and fusion part is removed. The results show that the early-fusion structure significantly outperforms others, showing an approximately 10\% improvement in prediction metrics compared with the non-conditional prediction. This clearly suggests that leveraging future information of the controllable agent (AV) could allow the prediction module to be more informed and the results more accurate. Nevertheless, the structure of fusing the AV's future information needs careful design. Here, we investigate three fusion structures: the early fusion structure treats the AV's future information as part of the query to the agent-map interaction encoder, the late fusion setup feeds the AV's future information only at the final decoding stage, and the early+late fusion structure uses both early and late fusion approaches. As the results show, the late-fusion or early-late-fusion variant performs significantly worse than the early-fusion structure and even worse than the non-conditional prediction network. This indicates that early fusion is a more effective structure in CMP modules and late fusion may influence the prediction results in a negative way.

\begin{figure*}[htp]
    \centering
    \includegraphics[width=0.755\linewidth]{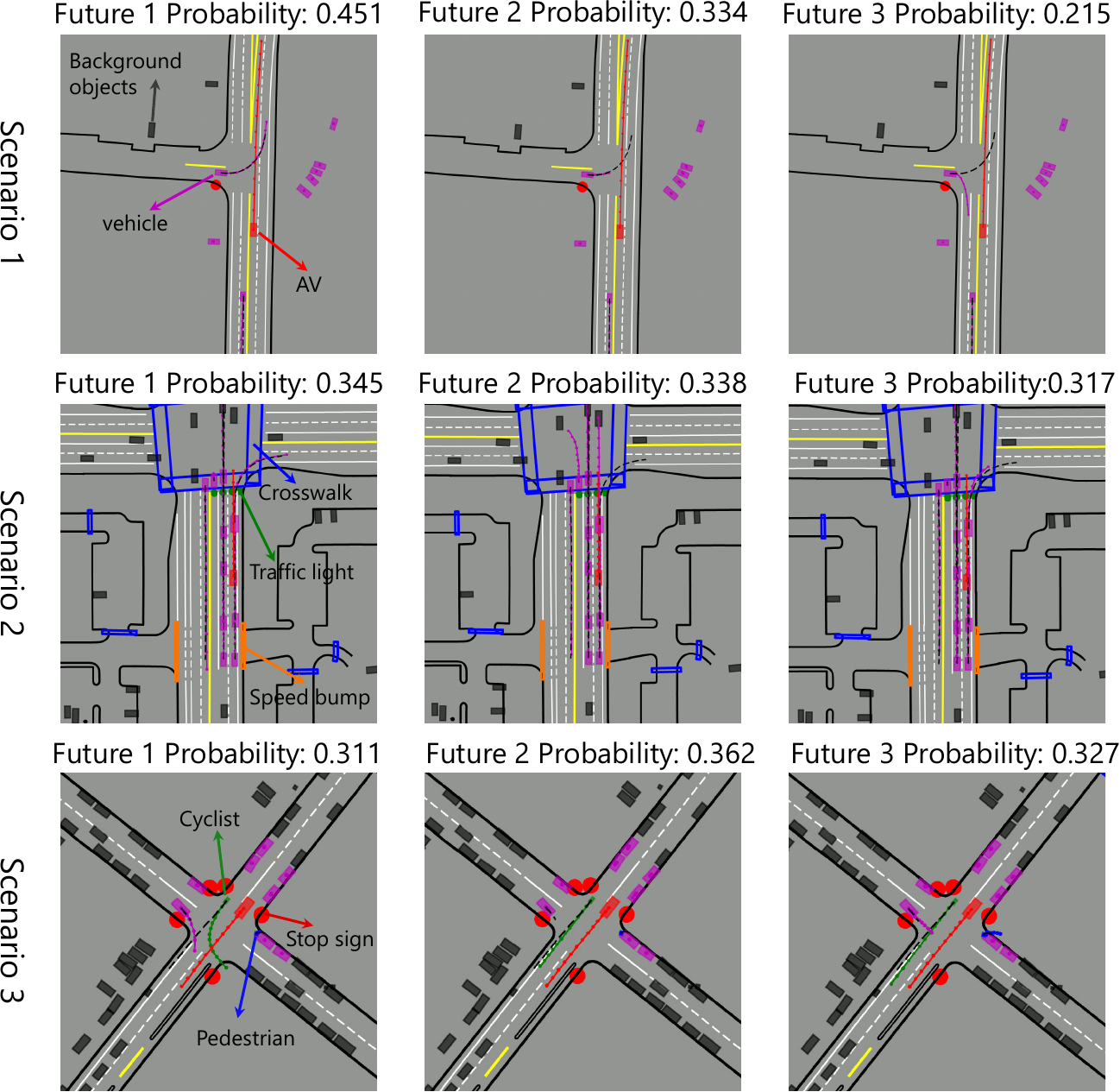}
    \caption{Qualitative results of the conditional prediction module to predict multiple futures for surrounding agents given a planned trajectory. The red box is the AV and other objects of interest are rendered in colored boxes; black boxes are background objects. Colored lines are predicted/planned trajectories and black dashed lines are ground-truth trajectories.}
    \label{fig:5}
\end{figure*}

\textbf{Multi-future prediction}.
Since the early fusion CMP module achieves the best prediction performance, we demonstrate the network's ability to jointly predict multiple futures for surrounding agents based on the early fusion structure. Fig. \ref{fig:5} shows three representative driving scenarios, each with three possible predicted futures given a single planned AV trajectory, and we use the ground-truth AV trajectory as the conditional information input to the CMP model. The prediction model can capture the multi-modality of agents' behaviors in accordance with the road structure and generate other agents' joint future trajectories in a scene-consistent manner (i.e., no self-collisions between predicted trajectories). The model can also assign a probability to each predicted future, and the predicted future closer to the ground-truth one is assigned with a higher probability. The results reveal the model's capability to handle the uncertainty of the future under the same future plan, thus enabling the downstream planner better evaluate the plan.

\begin{figure*}[htp]
    \centering
    \includegraphics[width=0.95\linewidth]{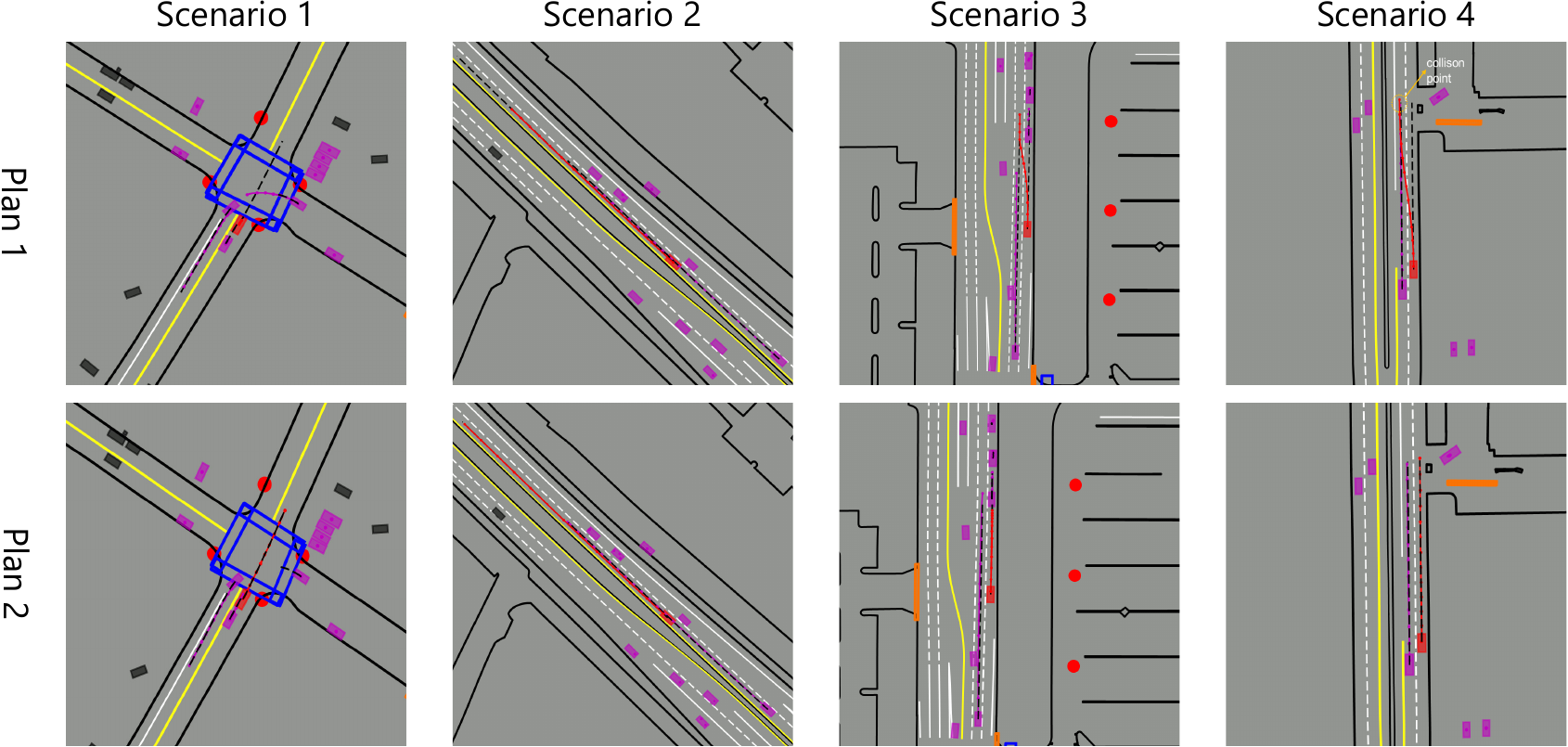}
    \caption{Qualitative results of the conditional prediction module to predict other agents' behaviors according to the AV's planned trajectory. Only the agents that have close interaction with the AV are highlighted and their trajectories are shown.}
    \label{fig:6}
\end{figure*}

\textbf{Conditional prediction}. 
We also adopt the early fusion structure as the prediction model and display the network's ability to predict other agents' behaviors conditioned on the AV's different plans in Fig. \ref{fig:6}. Note that only the most-likely prediction result given an AV's plan is shown for clarity. In Scenario 1, the AV is interacting with a vehicle that may conflict with the AV's route at an unsignalized intersection. Under the yield decision (Plan 1, target speed is 0), the other vehicle is predicted to pass first and the vehicle behind the AV is predicted to stay still. When giving a pass decision (Plan 2, target speed is high) to the AV, the other vehicle is predicted to yield and the vehicle behind starts moving. In Scenario 2, an expressway, we can see that if the AV chooses to decelerate (Plan 1), the vehicle behind the AV is predicted to decelerate too and keep a safe distance (compared to Plan 2). In Scenario 3, the other vehicle is predicted to slow down if the AV plans a lane change to avoid the congested lane (Plan 1) and keep the speed if the AV plans to decelerate in its current lane (Plan 2) without interfering with other vehicles. However, the model cannot completely make reactive predictions for other agents, which means other agents may not react to the AV's different plans and cause collisions in some plans, as shown in Scenario 4. This is because some of the AV's candidate plans deviate from the training data or normal behaviors and the model cannot make reactive predictions in such situations. Nonetheless, such plans will be ruled out by the downstream planner, which encourages the planner to choose plans that comply with the training distribution from real-world data.

\subsection{Planning performance}

\begin{figure*}[htp]
    \centering
    \includegraphics[width=0.95\linewidth]{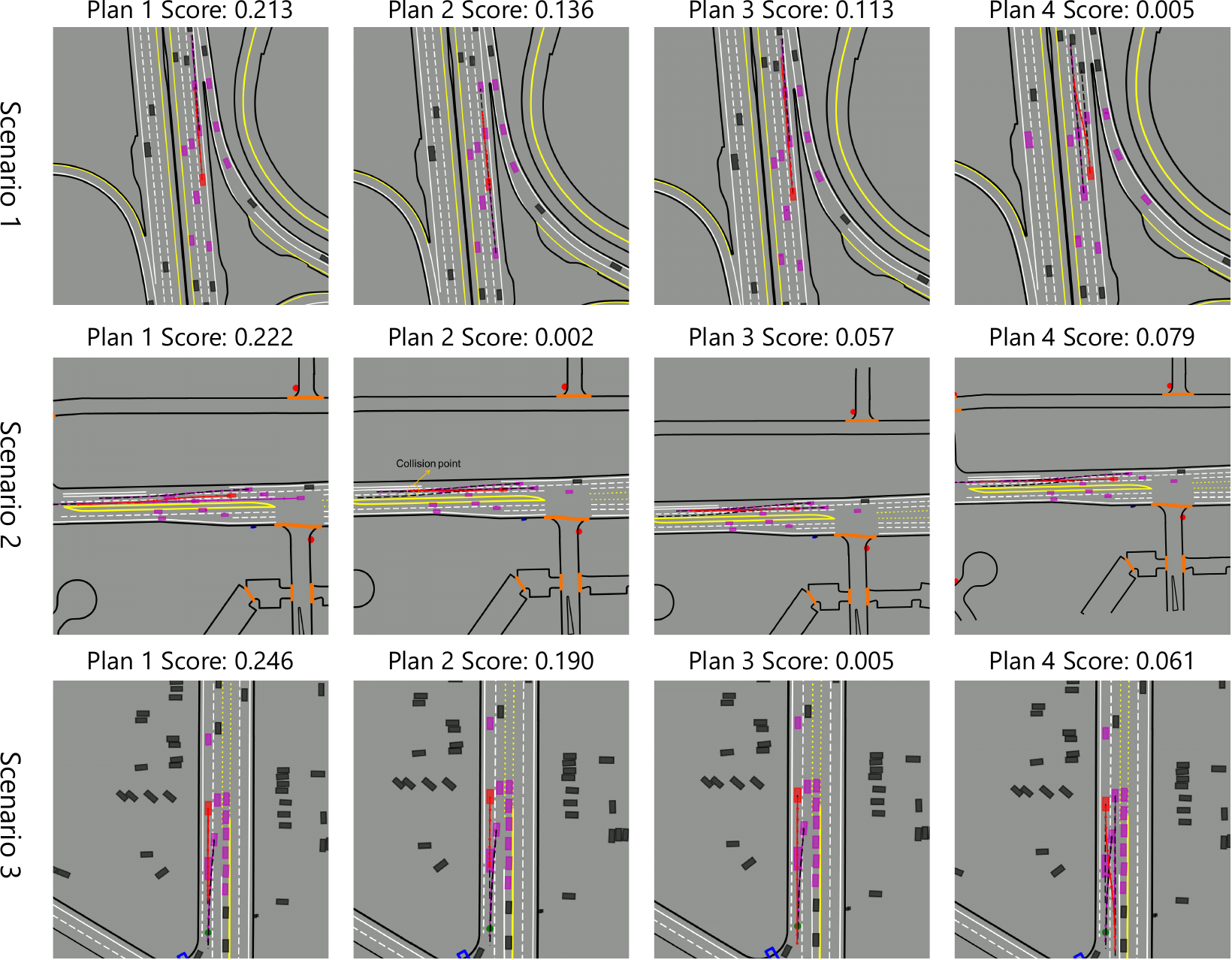}
    \caption{Qualitative results of the behavior planning module to properly score different candidate plans. Only the trajectories of the agents that could potentially influence the evaluation of the AV's plan are shown in each situation.}
    \label{fig:7}
\end{figure*}

\textbf{Qualitative results}. We utilize the testing set with $2,246$ urban driving scenes to evaluate the behavior planning performance of our method. We first display the qualitative results in Fig. \ref{fig:7}, showing the proposed method's capability to predict other agents' trajectories and select appropriate behaviors in some representative driving scenarios. For each scenario, we display four candidate behaviors and their normalized scores, and only the predicted future with the highest probability is shown for clarity. Scenario 1 shows a car-following scenario on a multi-lane expressway, where the AV should keep a safe distance from the leading vehicle. Plan 1, which is the closest to the ground truth, has the highest score among candidate trajectories. Other candidate plans (e.g., Plan 2 with lower target speed and Plan 3 with higher target speed) have lower scores because they would lead to smaller headway to the leading vehicle (unsafe behavior) or unnecessary speed loss. Changing lanes (Plan 4) is also unfavorable as it would induce unnecessary lateral discomfort without increasing the speed. In Scenario 2, the AV needs to deal with a cut-in vehicle from the right lane while also interacting with other vehicles. Our method selects the lane-changing behavior (Plan 1) with the highest score because it safely avoids the collision risk with the cut-in vehicle and also speed loss. However, if the AV tries hard braking to yield to the cut-in vehicle (Plan 2), there is a risk that the vehicle on the rear end could collide with the AV, and thus this plan has a near-to-zero score. Other candidate plans, such as slowing down without lane changing (Plan 3) and accelerating to overtake the cut-in vehicle (Plan 4), have low scores because they sacrifice the safety gap to the leading vehicle. In Scenario 3, slowing down and yielding to the cut-in vehicle (Plan 1 and Plan 2) have the higher scores because either accelerating (Plan 3) or changing lanes (Plan 4) could result in collision risks. The results suggest that the learned cost function can properly score different candidate plans according to the plan itself and also the conditional prediction result, and the behaviors (trajectory proposals) closest to ground truth can be assigned with the highest scores in most cases.

\textbf{Quantitative results}. To compare the model's planning performance, we set up several baseline methods and apply them to the behavior planning task in the testing scenes. \textbf{Neural network classifier}: we build a neural network that takes as input the planned trajectories and predicted trajectories from the CMP module and directly outputs the score of each planned trajectory. The neural network is trained with the classification (cross-entropy) loss and the same training data for planning. \textbf{Neural network regressor}: we build another neural network to directly output the target speed in the longitudinal direction and a value indicating lane change in the lateral direction. The network utilizes the backbone of the CMP module, which takes as input the information of the AV's and other agents' historical states and the map information. The network is trained with the mean squared error between the outputs and label speed and lane change indicator. \textbf{Model-based}: we use the intelligent driver model (IDM) to compute the desired speed and the minimizing overall braking induced by lane changes (MOBIL) algorithm to decide the lane-changing maneuver \cite{moghadam2021autonomous}. For the neural network regressor and model-based methods, a trajectory can be obtained given the target speed and lane change. We compare the obtained trajectories from different methods against the ground-truth trajectories in each scene and report the results in Table \ref{tab:2} considering the evaluation metrics previously defined.

\begin{table}[htp]
    \centering
    \caption{Evaluation of the planning performance in comparison with baseline methods}
    \resizebox{\linewidth}{!}{
    \begin{tabular}{@{}ccccc@{}}
    \toprule
                  & minFDE          & Accuracy (\%)     &  Speed Acc. (\%)  & Lane Acc. (\%)\\ \midrule
    Model-based   & 7.59            & --                &   75.15           & 75.63         \\
    NN regressor  & 5.66            & --                &   91.08           &  85.56        \\
    NN classifier & 2.91            & 68.61             & \textbf{96.59}    &  89.91        \\
    Ours          & \textbf{2.78}   & \textbf{69.88}    &   95.40           &  \textbf{90.12} \\ \bottomrule
    \end{tabular}
    }
    \label{tab:2}
\end{table}

The results in Table \ref{tab:2} reveal that our proposed method delivers human-like decision-making ability in terms of the position error to human driving trajectories and the accuracy of choosing closed-to-human behaviors. The performance of our method to make intention-level decisions is superior, reaching over 95\% of accuracy in target speed and over 90\% of accuracy in target lane compared to ground-truth human driving data. The NN classifier method, which can be regarded as deep IRL, achieves similar performance to our approach. However, the interpretability of such a method is compromised. On the other hand, the performance of the NN regressor method is inferior, which shows the drawback of learning-based methods that directly output decision values, lacking interpretability and robustness. The model-based method performs the worst because they are based on simple mathematical formulations and rules and is thereby not applicable to complex urban driving scenarios.

\subsection{Effects of the prediction model}
\begin{figure}[htp]
    \centering
    \includegraphics[width=\linewidth]{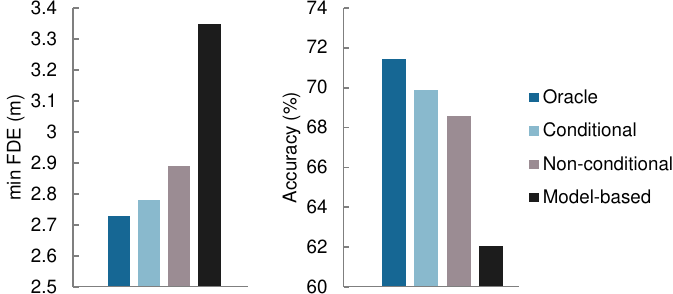}
    \caption{Evaluation of the influence of different prediction models on the planning performance}
    \label{fig:8}
    \vspace{-0.5cm}
\end{figure}

We investigate the influence of the prediction module on the downstream planning performance of our method. We utilize different prediction models in the behavior planning framework and test the planning performance in the same testing driving scenes. In addition to the proposed conditional prediction model (with early fusion structure), other prediction models used are listed as follows. \textbf{Non-conditional}: we remove the AV plan encoding and fusion parts from the proposed conditional prediction model. \textbf{Model-based}: we use the constant turn rate and velocity (CTRV) model to predict the surrounding agents' future trajectories and the model's prediction accuracy is very limited. \textbf{Oracle}: we use the ground-truth future trajectories as the predicted trajectories of surrounding agents to reveal the upper bound of the influence of prediction accuracy. Here, we report the primary planning evaluation metrics, i.e., minFDE and accuracy, and the results are given in Fig. \ref{fig:8}. 

The results indicate that the prediction accuracy plays an important role to ensure the downstream planning performance and using a learning-based prediction model can significantly improve the prediction accuracy and consequently planning performance compared to a kinematic-based prediction model. Moreover, using the conditional prediction model in the planning framework outperforms the non-conditional model, which stresses the benefit of leveraging the {AV's} future plan information in a prediction model. We also find that planning with the proposed conditional prediction model has comparable performance to the oracle method, which suggests that the conditional prediction model could better reflect the real-world interaction dynamics. There are two reasons why planning errors exist even when using the oracle model. First, the generated trajectory proposals are limited and coarse, and thus they cannot cover all the possible trajectories that a human driver may take. Another reason is the limitation of evaluation. Because the IRL module only uses a linear cost function, which cannot fully reflect a human {driver’s} actual evaluation of costs, the scoring of the generated trajectories may not be accurate in some cases.

\subsection{Effects of the cost function}
\begin{figure}[htp]
    \centering
    \includegraphics[width=\linewidth]{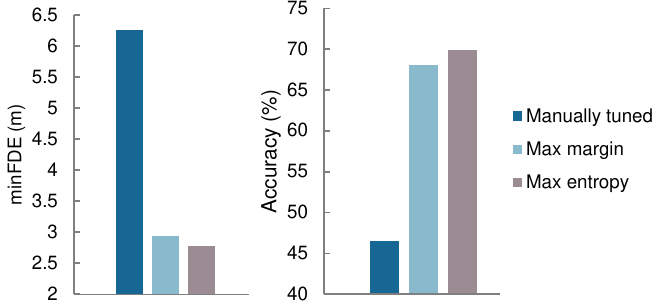}
    \caption{Evaluation of the influence of cost functions derived from different methods on the planning performance}
    \label{fig:9}
    \vspace{-0.5cm}
\end{figure}

We investigate the influence of the cost function to evaluate the candidate plans and the final planning performance. Here, we fix the prediction module as the proposed conditional prediction method and introduce two other baseline methods to obtain the cost function. \textbf{Manually tuned}: we manually tune the cost function weights according to a human expert's experience. \textbf{Maximum-margin}: we learn the cost function with the same training data using the max-margin method \cite{abbeel2004apprenticeship}, which is also a popular IRL algorithm. The results of planning performance with different cost functions derived from different methods are summarized in Fig. \ref{fig:9}. 

We can conclude that learning the cost function from data can significantly improve the evaluation of the candidate behaviors and human likeness compared to using a manually tuned cost function. In addition, the maximum-entropy IRL method and the maximum-margin IRL method have similar planning performance but the max-entropy method marginally outperforms the max-margin method. The results underscore the importance of learning the cost function from data rather than tuning it manually.

\subsection{Computation time}
We compare the computation time of different methods and all methods run on an NVIDIA RTX 3080 GPU. For the conditional prediction methods, there are two inference approaches: 1) single, which means we query the prediction model for each planned trajectory; 2) batch, which means we organize all the planned trajectories into a batch and repeat the environmental context tensors to match with the plan queries. The results in Table \ref{tab3} reveal that the batch processing method can significantly reduce the computation time by paralleling the conditional prediction process. The computation time can satisfy the real-time requirement as behavior planning runs at a lower frequency ($\leq$ 2 Hz). The single processing method that frequently queries the prediction model for multiple candidate plans has the longest computation time and is not suitable for real-time usage. The non-conditional method has the shortest computation time but the planning and prediction performance is a trade-off. In addition, the early fusion method runs slightly faster than the late fusion method.

\begin{table}[htp]
\centering
\caption{Comparison of computation time of different methods}
\begin{tabular}{@{}ccc@{}}
\toprule
Method                        & Inference & Time (ms) \\ \midrule
Non-conditional               & --        & 68         \\
\multirow{2}{*}{Early fusion} & Single    & 928        \\
                              & Batch     & 145        \\
\multirow{2}{*}{Late fusion}  & Single    & 963        \\
                              & Batch     & 149        \\ \bottomrule
\end{tabular}
\label{tab3}
\end{table}

We also report the computation time for learning the cost function using IRL. Note that IRL is only conducted offline, and the learned cost function is then directly used in online testing. The computational requirements for IRL are minimal, as only a few parameters in the cost function are learnable, but the computation of features may take up a large amount of time. Specifically, the computing time of a single IRL iteration (64 scenes in a batch) is approximately 28 seconds. Within this time, 9 $s$ are used to query the conditional prediction module for the responses of other agents, and the remaining 19 $s$ are spent on computing the feature vectors of different plans for all scenes across the batch. We iteratively feed different plans to the CMP model and use batch processing to obtain the prediction results and compute the features, so as to improve the computation efficiency. Nevertheless, the computational time is not a major concern in offline learning, and only a few hundred iterations are required to finish the learning process of the cost function weights. When using the scoring module in online testing, computing the feature vector of one candidate AV plan requires 120 $ms$. To meet the real-time requirement, we can parallelize the computing process to obtain the feature vectors of all candidate plans. Consequently, considering the computation time of generation, prediction, and scoring, our proposed framework can perform the behavior planning task within an acceptable time frame ($<$ 400 $ms$).

\subsection{Discussions}
The proposed framework divides the behavior planning task, one of the major challenges in autonomous driving, into prediction and scoring processes. Compared to other learning-based methods that typically only output the decision values, our framework that learns prediction and cost function to evaluate candidate plans can bring better safety, interpretability, and reliability to the system. Based on the overall framework, we propose a conditional motion prediction model that can forecast other agents' future trajectories according to the AV's potential future plan, which tightly couples the prediction and planning modules. The experiment results demonstrate that our framework has better planner performance (human-likeness or similarity to humans) compared to the neural network-based method that directly outputs decision values and the traditional model-based method. The proposed conditional prediction model and learning the cost function with inverse reinforcement learning both play important roles to ensure the proper and human-like evaluation of candidate plans. 

Nevertheless, some limitations of this work should be acknowledged. One limitation is that we only validate the framework in an open-loop manner because we ignore the low-level trajectory planner and controller. In our future work, we will test the proposed behavior planner in a closed-loop simulator to fully manifest its capabilities. Another limitation is that we do not investigate the performance of our framework in some safe-critical scenarios (e.g., encountering a road obstacle or complex intersection), and we plan to do that in our future work.

\section{Conclusions}
In this paper, we propose a learning-based predictive behavior planning framework that comprises three core modules: a behavior generation module, a conditional motion prediction module, and a scoring module. The behavior generation module produces a diverse set of trajectory proposals, while the conditional motion prediction module forecasts other agents' future trajectories jointly conditioned on each candidate plan. The scoring module evaluates the candidate plans using a cost function learned with maximum entropy inverse reinforcement learning (IRL). We conduct comprehensive experiments to validate the proposed framework on a large-scale real-world urban driving dataset. The qualitative results demonstrate that the conditional prediction module can predict multi-modal futures given a candidate plan and provide reactive predictions to different plans. Moreover, the scoring module, with the learned cost function, can properly select plans that are close to human-driving ones. The quantitative results suggest that early fusion is the most effective structure for the conditional prediction model. Additionally, we find that the conditional prediction model not only improves the prediction accuracy but also facilitates the downstream scoring module to better evaluate candidate decisions, thereby delivering human-like behaviors. Lastly, we note that learning the cost function is crucial in correctly evaluating the candidate plans to align with human values.



%





\ifCLASSOPTIONcaptionsoff
  \newpage
\fi





\bibliographystyle{IEEEtran}
\bibliography{Bibliography}

\vfill


\end{document}